\documentclass{article}

\usepackage{arxiv}

\usepackage[utf8]{inputenc} 
\usepackage[T1]{fontenc}    
\usepackage{hyperref}       
\usepackage{url}            
\usepackage{booktabs}       
\usepackage{amsfonts}       
\usepackage{nicefrac}       
\usepackage{microtype}      
\usepackage{lipsum}
\usepackage{graphicx}
\graphicspath{ {./images/} }
\usepackage{amsmath}
\usepackage{bm}

\title{From Black-Box to White-Box: Control-Theoretic Neural Network Interpretability}

\author{
 Jihoon Moon \\
  Department of Mechanical Engineering\\
  The Pennsylvania State University\\
  University Park, PA 16802 \\
  \texttt{jihoonmoon@psu.edu} \\
}

\begin{document}
\maketitle
\begin{abstract}
Deep neural networks achieve state-of-the-art performance but remain difficult to interpret mechanistically. In this work, we propose a control-theoretic framework that treats a trained neural network as a nonlinear state-space system and uses local linearization, controllability and observability Gramians, and Hankel singular values to analyze its internal computation. For a given input, we linearize the network around the corresponding hidden activation pattern and construct a state-space model whose state consists of hidden neuron activations. The input–state and state–output Jacobians define local controllability and observability Gramians, from which we compute Hankel singular values and associated modes. These quantities provide a principled notion of neuron and pathway importance: controllability measures how easily each neuron can be excited by input perturbations, observability measures how strongly each neuron influences the output, and Hankel singular values rank internal modes that carry input–output energy. We illustrate the framework on simple feedforward networks, including a 1–2–2–1 SwiGLU network and a 2–3–3–2 GELU network. By comparing different operating points, we show how activation saturation reduces controllability, shrinks the dominant Hankel singular value, and shifts the dominant internal mode to a different subset of neurons. The proposed method turns a neural network into a collection of local white-box dynamical models and suggests which internal directions are natural candidates for pruning or constraints to improve interpretability.
\end{abstract}


\section{Introduction}

\subsection{Objective}

Deep neural networks are now the dominant approach in areas such as computer vision, natural language processing, and control, yet we still have only a limited grasp of what they are doing internally \cite{zhang2021survey,fan2021interpretability}. In domains where safety, reliability, or scientific insight are important, it is not enough to observe high accuracy on benchmarks; we also need to understand \emph{how} a network processes information, which hidden units and pathways are actually doing the work, and how changes at the input propagate to the output \cite{guidotti2018survey,adadi2018peeking,lipton2018mythos,rudin2019stop}.

In this paper, we propose a control-theoretic perspective on this problem. We treat a trained feedforward neural network as a nonlinear state-space system and locally linearize it around a given operating point. The concatenated hidden activations are viewed as a state vector, and the Jacobians of the network define an effective (static) state-space model. From this model, we construct controllability and observability Gramians and a Hankel-like product whose eigenvalues and eigenvectors characterize dominant internal modes. The goal is to obtain a mechanistic, neuron-level interpretation of how input perturbations excite hidden units, how those units influence the output, and which internal directions are most important.

This work is not written from the perspective of a machine learning expert, but rather from that of a control engineer. Instead of proposing new training algorithms or architectures, our aim is to suggest a different way of \emph{looking} at trained networks, using tools that are standard in systems and control theory. We hope that this viewpoint can open a new direction for mechanistic interpretability and provide a simple, mathematically explicit starting point that other researchers can refine, extend, and adapt to more complex models such as large language models.

\subsection{Related Work on Neural Network Interpretability}

Neural network interpretability and explainable AI have been extensively studied in recent years, and several surveys organize the field into broad categories such as post-hoc explanations, intrinsically interpretable models, and representation-oriented analyses \cite{zhang2021survey,fan2021interpretability,guidotti2018survey,adadi2018peeking}. These works emphasize that modern deep networks are typically used as black boxes, and that there is a tension between predictive performance, interpretability, and the demands of high-stakes decision-making \cite{lipton2018mythos,rudin2019stop}.

A large class of methods focuses on \emph{post-hoc} explanations of individual predictions. Saliency-based techniques highlight important input features by computing or approximating gradients of the output with respect to the input \cite{simonyan2013deep}. Local surrogate models and perturbation-based methods, such as LIME, fit simple interpretable models around a given input to explain the behavior of a complex classifier \cite{ribeiro2016should}. Attribution methods like Integrated Gradients and Shapley-value-based approaches provide more principled feature-importance scores by enforcing axioms or game-theoretic properties \cite{sundararajan2017axiomatic,lundberg2017unified}. These methods explain \emph{what} parts of the input matter for a prediction, but typically do not provide a mechanistic account of the internal computation across layers.

More recently, \emph{mechanistic interpretability} has emerged as a line of work that attempts to reverse engineer neural networks at the level of their internal structure and computation, going beyond input–output explanations. This includes neuron and feature visualization, analysis of intermediate representations, and circuit-level studies of how subnetworks implement specific behaviors \cite{olah2018building,olah2020zoom}. In the context of transformers and large language models, researchers have developed mathematical frameworks for transformer circuits and performed detailed case studies that connect particular attention heads, multilayer perceptron neurons, and residual streams to interpretable algorithms or stored information \cite{elhage2021transformercircuits}.

A particularly relevant recent contribution is the work on weight-sparse transformers with interpretable circuits \cite{gao2025weightsparse}. There, transformers are trained with strong sparsity constraints so that most weights become exactly zero, and additional pruning isolates small task-specific circuits. The authors show that on carefully designed code-understanding tasks, the resulting circuits are compact and many nodes correspond to natural concepts, and they validate these circuits by ablation experiments that preserve or destroy performance. This provides some of the clearest circuit-level explanations to date. However, the approach requires training specialized sparse models, is demonstrated on relatively small scales and narrow tasks, and does not directly apply to standard dense networks used in practice. This motivates complementary approaches that work directly with existing models and do not rely on sparsity-inducing training.

In parallel, there is a line of work that explicitly models neural networks or neural network layers as dynamical systems, particularly in the form of deep state-space models for long-range sequence modeling \cite{gu2021efficiently}. In classical control theory, controllability and observability Gramians, Hankel operators, and balanced truncation provide powerful tools for analyzing and reducing linear dynamical systems, with Hankel singular values quantifying how much each internal mode contributes to the input–output map \cite{moore2003principal,antoulas2005approximation,khalil1996robust}. Recent work has begun to bring these ideas into modern machine learning, for example by using Hankel singular values as regularizers for compressible state-space models \cite{schwerdtner2025hankel}. Despite this, relatively little work explicitly connects Gramian-based notions of controllability and observability to neuron-level interpretability in standard feedforward networks.

Most existing interpretability methods either operate directly on the input space, learn a separate explanatory model, or treat hidden units as black boxes to be visualized or probed. By contrast, the present work imports tools from systems and control theory to quantify how hidden states can be influenced by the input (controllability), how they influence the output (observability), and how these properties combine into dominant internal modes ranked by Hankel singular values. Our framework is intended as a simple bridge between these two communities: it uses standard Jacobians and matrix operations available in modern machine learning tool chains, but interprets them through the lens of classical system-theoretic concepts.

\subsection{Contributions}

This paper makes the following contributions from a control-theoretic perspective:

\begin{itemize}
  \item We formulate a trained feedforward neural network as a nonlinear state-space system whose state is the stacked hidden activation vector. Around a chosen operating point, we derive the input–state Jacobian and the hidden–output Jacobian via standard differentiation.
  \item We define static analogues of the controllability and observability Gramians, and interpret their diagonal entries as per-neuron measures of how easily each hidden unit is excited by the input and how strongly it affects the output, respectively.
  \item We construct a Hankel-like product and analyze its eigenvalues and eigenvectors, obtaining Hankel singular values and associated internal modes. These modes define directions in hidden-state space that are jointly important for controllability and observability, in a manner analogous to classical balanced truncation \cite{moore2003principal,antoulas2005approximation,khalil1996robust}.
  \item We propose a neuron-level importance metric that aggregates each neuron's participation across modes, and we illustrate the resulting mechanistic interpretation on two small example networks with SwiGLU and GELU activations. The examples demonstrate that the framework recovers a low-dimensional internal pathway and highlights a small subset of neurons as locally dominant, while also revealing how activation saturation and operating point affect controllability, observability, and the dominant modes.
\end{itemize}

The goal is not to propose a complete theory of neural network interpretability, but to introduce a simple and extensible framework that uses familiar tools from control theory to shed light on the internal computation of trained networks. We hope this perspective can inspire further work that combines system-theoretic analysis with modern deep learning, including applications to larger architectures such as transformers and large language models.

\section{Methodology}\label{sec:methodology}

\subsection{Problem Formulation}
Let $f_\theta:\mathbb{R}^{n_x}\to\mathbb{R}^{n_y}$ be a trained neural network with parameters $\theta$. Our goal is to obtain a \emph{mechanistic, dynamical} description of how internal neurons mediate the mapping from input $x$ to output $y=f_\theta(x)$, and to quantify the importance of individual neurons and internal directions.

We pursue three main steps:
\begin{enumerate}
  \item Represent the network as a nonlinear state-space system whose state is the vector of hidden activations;
  \item Perform a local linearization of this system around a given operating point $x^*$;
  \item Use controllability and observability Gramians and Hankel singular values of the resulting linear model to identify dominant internal modes and neuron importance.
\end{enumerate}

This yields a local \emph{white-box} approximation of the network around $x^*$, which can be repeated at different inputs to study input-dependent mechanisms.

\subsection{Network Representation and State Definition}
We consider a feedforward network with $L-1$ hidden layers and an output layer. Let
\[
  n_0 = n_x, \qquad n_L = n_y,
\]
and let layer $\ell$ have width $n_\ell$ for $\ell=1,\dots,L-1$. The forward map is
\begin{align}
  h^{(0)} &= x \in \mathbb{R}^{n_0},\\
  z^{(\ell)} &= W_\ell h^{(\ell-1)} + b_\ell, \qquad W_\ell \in \mathbb{R}^{n_\ell \times n_{\ell-1}},\\
  h^{(\ell)} &= \bm{\sigma}_\ell\big(z^{(\ell)}\big) \in \mathbb{R}^{n_\ell}, \qquad \ell = 1,\dots,L-1,\\
  z^{(L)} &= W_L h^{(L-1)} + b_L, \qquad W_L \in \mathbb{R}^{n_L \times n_{L-1}},\\
  y &= \bm{\sigma}_L\big(z^{(L)}\big) \in \mathbb{R}^{n_y},
\end{align}
where each $\bm{\sigma}_\ell$ is activation function (e.g., SwiGLU, GELU, ReLU).

We define the hidden state vector as the concatenation of all hidden activations:
\begin{equation}
  h = \operatorname{col}\big(h^{(1)},\dots,h^{(L-1)}\big) \in \mathbb{R}^{n_h},
  \qquad n_h = \sum_{\ell=1}^{L-1} n_\ell.
\end{equation}

Our objective is to derive a linear approximation
\begin{equation}
  \delta h \approx B\,\delta x, \qquad \delta y \approx C \delta h,
\end{equation}
around a fixed operating point $(x^*, h^*, y^*)$, and then analyze this linear model using tools from linear systems theory.

\subsection{Local Linearization}

\subsubsection{Operating Point and Perturbations}
Fix an input $x^*$ and compute the corresponding forward pass:
\begin{align}
  z^{(\ell)*} &= W_\ell h^{(\ell-1)*} + b_\ell,\\
  h^{(\ell)*} &= \bm{\sigma}_\ell\big(z^{(\ell)*}\big), \qquad \ell = 1,\dots,L-1,\\
  z^{(L)*} &= W_L h^{(L-1)*} + b_L,\\
  y^* &= \bm{\sigma}_L\big(z^{(L)*}\big).
\end{align}

We define perturbations around this point as
\begin{equation}
  \delta x = x - x^*, \qquad
  \delta h^{(\ell)} = h^{(\ell)} - h^{(\ell)*}, \qquad
  \delta y = y - y^*.
\end{equation}

\subsubsection{Layer Jacobians}
For each hidden layer $\ell$, define the diagonal activation derivative matrix
\begin{equation}
  D_\ell := \operatorname{diag}\big(\bm{\sigma}_\ell'(z^{(\ell)*})\big) \in \mathbb{R}^{n_\ell \times n_\ell}.
\end{equation}
Using the chain rule,
\begin{equation}
  \frac{\partial h^{(\ell)}}{\partial z^{(\ell)}}\Big|_{x^*} = D_\ell, \qquad
  \frac{\partial z^{(\ell)}}{\partial h^{(\ell-1)}} = W_\ell,
\end{equation}
so the Jacobian of layer $\ell$ with respect to its input is
\begin{equation}
  J_\ell := \frac{\partial h^{(\ell)}}{\partial h^{(\ell-1)}}\Big|_{x^*}
  = D_\ell W_\ell \in \mathbb{R}^{n_\ell \times n_{\ell-1}}.
\end{equation}
Thus small perturbations propagate layer-by-layer as
\begin{equation}
  \delta h^{(\ell)} \approx J_\ell \, \delta h^{(\ell-1)}.
\end{equation}

For the output layer, define
\begin{equation}
  D_L := \operatorname{diag}\big(\bm{\sigma}_L'(z^{(L)*})\big) \in \mathbb{R}^{n_y \times n_y},
\end{equation}
then the Jacobian of the output with respect to the last hidden layer is
\begin{equation}
  C_{\text{local}} := \frac{\partial y}{\partial h^{(L-1)}}\Big|_{x^*}
  = D_L W_L \in \mathbb{R}^{n_y \times n_{L-1}}.
\end{equation}
If $\bm{\sigma}_L$ is linear, then $D_L = I$ and $C_{\text{local}} = W_L$.

\subsubsection{Global Input--Hidden Jacobian $B$}
The multivariate chain rule gives, for each hidden layer $\ell$,
\begin{equation}
  \frac{\partial h^{(\ell)}}{\partial x}\Big|_{x^*}
  = \frac{\partial h^{(\ell)}}{\partial h^{(\ell-1)}}
    \frac{\partial h^{(\ell-1)}}{\partial h^{(\ell-2)}} \cdots
    \frac{\partial h^{(1)}}{\partial x}
  = J_\ell J_{\ell-1} \cdots J_1.
\end{equation}
We define the block input--hidden Jacobian for layer $\ell$ as
\begin{equation}
  B^{(\ell)} := \frac{\partial h^{(\ell)}}{\partial x}\Big|_{x^*}
  = J_\ell J_{\ell-1} \cdots J_1
  = (D_\ell W_\ell)(D_{\ell-1} W_{\ell-1}) \cdots (D_1 W_1),
\end{equation}
where the order of the matrix product is crucial and follows the chain rule.

Stacking all hidden layers, we obtain the full input--state Jacobian
\begin{equation}
  B =
  \begin{bmatrix}
    B^{(1)}\\
    B^{(2)}\\
    \vdots\\
    B^{(L-1)}
  \end{bmatrix}
  \in \mathbb{R}^{n_h \times n_x},
\end{equation}
where each row block $B^{(\ell)} \in \mathbb{R}^{n_\ell \times n_x}$ captures how the $\ell$-th hidden layer responds to input perturbations near $x^*$.

Elementwise,
\begin{equation}
  B_{ij} = \frac{\partial h_i}{\partial x_j}\Big|_{x^*},
\end{equation}
where $h_i$ is the $i$-th component of the stacked hidden state.

\subsubsection{Hidden--Output Jacobian $C$}
The output $y$ depends only on the last hidden layer $h^{(L-1)}$. In terms of the stacked state $h$, the hidden--output Jacobian is
\begin{equation}
  C =
  \begin{bmatrix}
    0 & \cdots & 0 & C_{\text{local}}
  \end{bmatrix}
  \in \mathbb{R}^{n_y \times n_h},
\end{equation}
where the nonzero block $C_{\text{local}}$ occupies the columns corresponding to $h^{(L-1)}$, and earlier hidden layers have zero direct influence.

Collecting these results, the local linear approximation of the nonlinear network around $x^*$ is
\begin{equation}
  \delta h = B\,\delta x, \qquad \delta y = C \, \delta h.
\end{equation}

\subsection{Gramian Construction}
We now use the linearized model to define controllability and observability Gramians for the hidden state. Although the model is static (no explicit recurrence in depth), these Gramians provide meaningful measures of how easily each neuron can be influenced by the input and how strongly it influences the output.

\subsubsection{Controllability Gramian}
For a discrete-time linear system
\begin{equation}
    h_{k+1} = A h_k + B u_k,
\end{equation}
the (infinite-horizon) controllability Gramian $W_C$ is defined as the unique positive semidefinite solution of the discrete Lyapunov equation
\begin{equation}
    W_C = A W_C A^\top + B B^\top,
\end{equation}
In our setting, the network is locally approximated by the static map
\(
\delta h = B\,\delta x
\),
with no internal recurrence. This corresponds to the special case
\(
A = 0
\)
in the above state-space model. Substituting $A=0$ into the Lyapunov equation gives
\begin{equation}
    W_C = 0 \cdot W_C \cdot 0^\top + B B^\top = B B^\top,
\end{equation}
so that the controllability Gramian for our locally linearized network reduces to
\begin{equation}
    W_C := B B^\top \in \mathbb{R}^{n_h \times n_h}.
\end{equation}
Thus each diagonal entry $(W_C)_{ii}$ measures the squared sensitivity of the $i$-th hidden state component to perturbations in the input.
The diagonal entries
\begin{equation}
  (W_C)_{ii} = \left\lVert \frac{\partial h_i}{\partial x}\Big|_{x^*} \right\rVert_2^2
\end{equation}
measure how sensitive neuron $i$ is to small perturbations in the input at $x^*$. Large values indicate neurons that are easily controllable; small values indicate neurons that are locally difficult to excite.

\subsubsection{Observability Gramian}
Similarly, we define the observability Gramian as
\begin{equation}
  W_O := (C)^\top C \in \mathbb{R}^{n_h \times n_h}.
\end{equation}
The diagonal entries
\begin{equation}
  (W_O)_{ii} = \left\lVert \frac{\partial y}{\partial h_i}\Big|_{x^*} \right\rVert_2^2
\end{equation}
measure how strongly neuron $i$ influences the output. Large values indicate neurons that are highly observable; small values indicate neurons whose perturbations have little local effect on the output. Off-diagonal entries capture correlations in how neurons are jointly driven or observed.

\subsection{Hankel Singular Values and Mode Analysis}
To jointly capture controllability and observability, we consider the product
\begin{equation}
  M := W_C W_O \in \mathbb{R}^{n_h \times n_h}.
\end{equation}
Let $\lambda_i(M)$ denote its eigenvalues and let $v_i \in \mathbb{R}^{n_h}$ be associated eigenvectors. We define the Hankel singular values as
\begin{equation}
  \sigma_i := \sqrt{\lambda_i(M)}, \qquad i = 1,\dots,r, \quad r \le n_h,
\end{equation}
and refer to the pairs $(\sigma_i, v_i)$ as the internal modes of the linearized network at $x^*$.

Each mode $v_i$ defines a direction in the hidden state space
\begin{equation}
  z^{(i)} = v_i^\top h,
\end{equation}
and $\sigma_i$ quantifies how strongly this direction participates in the input--output map: modes with large $\sigma_i$ are both easily excited by the input (high controllability) and strongly visible at the output (high observability). Modes with very small $\sigma_i$ are locally negligible and can be considered candidates for pruning or reduction.

\subsubsection{Neuron-Level Importance}
The contribution of neuron $j$ to mode $i$ is captured by the squared component $v_{i,j}^2$. We define an overall importance score for neuron $j$ by aggregating across modes:
\begin{equation}
  \mathrm{Imp}(j) := \sum_{i=1}^{r} \sigma_i^\alpha v_{i,j}^2,
\end{equation}
with $\alpha \ge 1$ (e.g., $\alpha = 1$ or $2$) controlling how strongly we emphasize higher-energy modes.

Neurons with large $\mathrm{Imp}(j)$ lie in one or more dominant modes and are thus mechanistically important near $x^*$. Neurons with very small $\mathrm{Imp}(j)$ participate only in low-energy modes and are locally unimportant to the input--output behavior.

\subsection{Algorithmic Summary}
For a given trained network $f_\theta$ and operating point $x^*$, the analysis proceeds as follows:
\begin{enumerate}
  \item \textbf{Forward pass:} compute $z^{(\ell)*}, h^{(\ell)*}, z^{(L)*}, y^*$.
  \item \textbf{Activation derivatives:} form $D_\ell = \operatorname{diag}(\bm{\sigma}_\ell'(z^{(\ell)*}))$ for $\ell = 1,\dots,L$.
  \item \textbf{Layer Jacobians:} compute $J_\ell = D_\ell W_\ell$ for $\ell = 1,\dots,L-1$ and $C_{\text{local}} = D_L W_L$.
  \item \textbf{Global Jacobians:} compute block Jacobians $B^{(\ell)} = J_\ell \cdots J_1$, stack to form $B$, and construct $C$.
  \item \textbf{Gramians:} form $W_C = B B^\top$ and $W_O = (C)^\top C$.
  \item \textbf{Modes:} compute eigenpairs $(\lambda_i, v_i)$ of $M = W_C W_O$, set $\sigma_i = \sqrt{\lambda_i}$.
  \item \textbf{Interpretation:} compute neuron importance $\mathrm{Imp}(j)$, visualize dominant modes, and analyze how internal pathways support the input--output behavior.
\end{enumerate}

By repeating this procedure at different inputs $x^*$, we obtain a family of local linear models that collectively describe how the network's internal computation and neuron importance change across the input space.

\section{Result}
\subsection{Example~1: $1$--$2$--$2$--$1$ Network with SwiGLU Activation}

We first consider a scalar-input, scalar-output network with two hidden layers. More precisely,
\begin{itemize}
  \item $n_0 = 1$ (input layer),
  \item $n_1 = 2$ (first hidden layer),
  \item $n_2 = 2$ (second hidden layer),
  \item $n_3 = 1$ (output layer), with $n_L = n_3 = n_y = 1$.
\end{itemize}

At layer $\ell \in \{1,2,3\}$ we use a SwiGLU activation $\bm{\sigma}_\ell(\cdot)$, applied to the pre-activation $z^{(\ell)}$. For our analysis we only require that $\bm{\sigma}_\ell$ be differentiable; we denote its derivative by $\bm{\sigma}'_\ell$ and obtain it via automatic differentiation.

The forward map is
\begin{align}
  h^{(0)} &= x \in \mathbb{R},\\[2pt]
  z^{(1)} &= W_1 h^{(0)} + b_1, & h^{(1)} &= \bm{\sigma}_1\big(z^{(1)}\big) \in \mathbb{R}^2, \\[2pt]
  z^{(2)} &= W_2 h^{(1)} + b_2, & h^{(2)} &= \bm{\sigma}_2\big(z^{(2)}\big) \in \mathbb{R}^2, \\[2pt]
  z^{(3)} &= W_3 h^{(2)} + b_3, & y &= h^{(3)} = \bm{\sigma}_3\big(z^{(3)}\big) \in \mathbb{R},
\end{align}
where $W_1 \in \mathbb{R}^{2\times 1}$, $W_2 \in \mathbb{R}^{2\times 2}$ and $W_3 \in \mathbb{R}^{1\times 2}$.

We define the stacked hidden state
\begin{equation}
  h = \operatorname{col}\big(h^{(1)}, h^{(2)}\big) \in \mathbb{R}^{n_h}, \qquad n_h = 4.
\end{equation}
For a given operating point $x^*$ we perform a forward pass to obtain $z^{(\ell)*}$ and $h^{(\ell)*}$ for $\ell=1,2,3$, and compute the diagonal derivative matrices
\begin{equation}
  D_\ell = \operatorname{diag}\big(\bm{\sigma}'_\ell(z^{(\ell)*})\big), \qquad \ell = 1,2,3.
\end{equation}
The layer Jacobians are
\begin{equation}
  J_1 = D_1 W_1 \in \mathbb{R}^{2\times 1},\qquad
  J_2 = D_2 W_2 \in \mathbb{R}^{2\times 2},\qquad
  C_{\text{local}} = D_3 W_3 \in \mathbb{R}^{1\times 2}.
\end{equation}
Using the general formulation of Section~\ref{sec:methodology}, the block input--hidden Jacobians are
\begin{equation}
  B^{(1)} = J_1 = D_1 W_1 \in \mathbb{R}^{2\times 1}, \qquad
  B^{(2)} = J_2 J_1 = (D_2 W_2)(D_1 W_1) \in \mathbb{R}^{2\times 1},
\end{equation}
and the full input--state Jacobian is
\begin{equation}
  B = \begin{bmatrix} B^{(1)} \\[2pt] B^{(2)} \end{bmatrix} \in \mathbb{R}^{4\times 1}.
\end{equation}
In terms of the stacked state $h = \operatorname{col}(h^{(1)}_1, h^{(1)}_2, h^{(2)}_1, h^{(2)}_2)$, the hidden--output Jacobian is
\begin{equation}
  C = \begin{bmatrix} 0 & 0 & C_{\text{local}} \end{bmatrix} \in \mathbb{R}^{1\times 4},
\end{equation}
so only the second hidden layer is directly observable from the output.

\paragraph{Numeric illustration.}
To make these quantities concrete, we fix specific weights and biases for the $1$--$2$--$2$--$1$ network and evaluate the Gramians at a single operating point $x^* = 0.5$. We choose
\begin{align}
  W_1 &= \begin{bmatrix} 1.0 \\ -0.5 \end{bmatrix}, & b_1 &= \begin{bmatrix} 0.0 \\ 0.2 \end{bmatrix},\\[4pt]
  W_2 &= \begin{bmatrix} 0.8 & -0.4 \\ 0.5 & 0.9 \end{bmatrix}, & b_2 &= \begin{bmatrix} 0.1 \\ -0.2 \end{bmatrix},\\[4pt]
  W_3 &= \begin{bmatrix} 1.1 & -0.7 \end{bmatrix}, & b_3 &= \begin{bmatrix} 0.0 \end{bmatrix}.
\end{align}
At $x^* = 0.5$, the forward pass yields hidden activations $h^{(1)*} \in \mathbb{R}^2$, $h^{(2)*} \in \mathbb{R}^2$ and output $y^* \in \mathbb{R}$. Using the SwiGLU derivatives at this point, the resulting input--state Jacobian is
\begin{equation}
  B \approx
  \begin{bmatrix}
    0.740 \\
   -0.238 \\
    0.464 \\
    0.073
  \end{bmatrix} \in \mathbb{R}^{4\times 1}.
\end{equation}
The controllability Gramian $W_C = B B^\top$ therefore has diagonal entries
\begin{equation}
  \operatorname{diag}(W_C) \approx [0.548,\; 0.056,\; 0.215,\; 0.005],
\end{equation}
indicating that the first neuron in the first hidden layer is the most controllable, followed by the first neuron in the second hidden layer, whereas the second neuron in the second layer is almost uncontrollable at this input.

The hidden--output Jacobian in stacked coordinates is
\begin{equation}
  C \approx \begin{bmatrix} 0 & 0 & 0.689 & -0.438 \end{bmatrix},
\end{equation}
so the observability Gramian $W_O = (C)^\top C$ has
\begin{equation}
  \operatorname{diag}(W_O) \approx [0.000,\; 0.000,\; 0.474,\; 0.192].
\end{equation}
As expected, only the second hidden layer is directly observable, with neuron $h^{(2)}_1$ having the strongest direct influence on the output.

The product $M = W_C W_O$ has a single nonzero eigenvalue
\begin{equation}
  \lambda_1 \approx 8.27\times 10^{-2},\qquad \sigma_1 = \sqrt{\lambda_1} \approx 0.288,
\end{equation}
with associated normalized eigenvector
\begin{equation}
  v_1 \approx [\,0.815,\; -0.262,\; 0.511,\; 0.080\,]^\top.
\end{equation}
This dominant Hankel mode corresponds to the scalar internal coordinate $z^{(1)} = v_1^\top h$, which can be interpreted as the primary internal pathway that carries input--output energy at $x^*$.

Using the importance metric $\mathrm{Imp}(j) = \sigma_1 v_{1,j}^2$ for $j=1,\dots,4$, we obtain
\begin{equation}
  \mathrm{Imp} \approx [0.191,\; 0.020,\; 0.075,\; 0.002].
\end{equation}
Thus the first hidden neuron $h^{(1)}_1$ dominates the mode, the first neuron in the second layer $h^{(2)}_1$ has secondary importance, the second neuron in the first layer plays a smaller but non-negligible role, and the second neuron in the second layer is almost irrelevant to the dominant internal pathway. This ranking is consistent with both the controllability and observability profiles: $h^{(1)}_1$ is highly controllable, and $h^{(2)}_1$ is highly observable.

\subsection{Example~2: $2$--$3$--$3$--$2$ Network with GELU Activation}

Our second example considers a higher-dimensional setting with multiple inputs and outputs. Details of number of neurons and hidden layers are
\begin{itemize}
  \item $n_0 = n_x = 2$ (input layer),
  \item $n_1 = 3$ (first hidden layer),
  \item $n_2 = 3$ (second hidden layer),
  \item $n_3 = n_L = n_y = 2$ (output layer).
\end{itemize}
We use the GELU nonlinearity at all hidden and output layers. The GELU activation $\bm{\sigma}_\ell(\cdot)$ with derivative $\bm{\sigma}'_\ell(z)$ computed analytically or via automatic differentiation and used to form the diagonal matrices $D_\ell$.

The forward map is
\begin{align}
  h^{(0)} &= x \in \mathbb{R}^2,\\[2pt]
  z^{(1)} &= W_1 h^{(0)} + b_1, & h^{(1)} &= \bm{\sigma}_1\big(z^{(1)}\big) \in \mathbb{R}^3, \\[2pt]
  z^{(2)} &= W_2 h^{(1)} + b_2, & h^{(2)} &= \bm{\sigma}_2\big(z^{(2)}\big) \in \mathbb{R}^3, \\[2pt]
  z^{(3)} &= W_3 h^{(2)} + b_3, & y &= h^{(3)} = \bm{\sigma}_3\big(z^{(3)}\big) \in \mathbb{R}^2,
\end{align}
where $W_1 \in \mathbb{R}^{3\times 2}$, $W_2 \in \mathbb{R}^{3\times 3}$ and $W_3 \in \mathbb{R}^{2\times 3}$.

We define the stacked hidden state
\begin{equation}
  h = \operatorname{col}\big(h^{(1)}, h^{(2)}\big) \in \mathbb{R}^{n_h}, \qquad n_h = 6.
\end{equation}
For a chosen operating point $x^* \in \mathbb{R}^2$ we compute $z^{(\ell)*}$ and $h^{(\ell)*}$ and form the diagonal matrices
\begin{equation}
  D_\ell = \operatorname{diag}\big(\bm{\sigma}'_l(z^{(\ell)*})\big), \qquad \ell = 1,2,3.
\end{equation}
The layer Jacobians are
\begin{equation}
  J_1 = D_1 W_1 \in \mathbb{R}^{3\times 2}, \qquad
  J_2 = D_2 W_2 \in \mathbb{R}^{3\times 3}, \qquad
  C_{\text{local}} = D_3 W_3 \in \mathbb{R}^{2\times 3}.
\end{equation}
Using the ordered products of Section~\ref{sec:methodology}, the block input--hidden Jacobians are
\begin{equation}
  B^{(1)} = J_1 = D_1 W_1 \in \mathbb{R}^{3\times 2}, \qquad
  B^{(2)} = J_2 J_1 = (D_2 W_2)(D_1 W_1) \in \mathbb{R}^{3\times 2},
\end{equation}
so that the full input--state Jacobian is
\begin{equation}
  B = \begin{bmatrix} B^{(1)} \\[2pt] B^{(2)} \end{bmatrix} \in \mathbb{R}^{6\times 2}.
\end{equation}
In terms of the stacked state
\[
  h = \operatorname{col}(h^{(1)}_1, h^{(1)}_2, h^{(1)}_3, h^{(2)}_1, h^{(2)}_2, h^{(2)}_3),
\]
the hidden--output Jacobian is
\begin{equation}
  C = \begin{bmatrix} 0 & 0 & 0 & C_{\text{local}} \end{bmatrix} \in \mathbb{R}^{2\times 6},
\end{equation}
so only the second hidden layer is directly observable from the outputs.

\paragraph{Numeric illustration.}
To instantiate these quantities, we fix specific trained weights and biases for the $2$--$3$--$3$--$2$ GELU network and evaluate the Gramians at an operating point $x^* = [0.3,\; -0.2]^\top$. We choose
\begin{align}
  W_1 &= \begin{bmatrix}
           0.8 & -0.3 \\
           0.5 &  0.7 \\
          -0.2 &  1.0
         \end{bmatrix}, &
  b_1 &= \begin{bmatrix} 0.1 \\ -0.1 \\ 0.0 \end{bmatrix}, \\[4pt]
  W_2 &= \begin{bmatrix}
           0.9 & -0.4 & 0.2 \\
           0.3 &  0.8 & -0.5 \\
          -0.6 &  0.1 & 1.0
         \end{bmatrix}, &
  b_2 &= \begin{bmatrix} 0.0 \\ 0.05 \\ -0.1 \end{bmatrix}, \\[4pt]
  W_3 &= \begin{bmatrix}
           1.0 & -0.5 & 0.3 \\
          -0.7 &  0.2 & 0.9
         \end{bmatrix}, &
  b_3 &= \begin{bmatrix} 0.0 \\ 0.1 \end{bmatrix}.
\end{align}
Evaluating the Jacobians at $x^*$ yields the input--state Jacobian
\begin{equation}
  B \approx
  \begin{bmatrix}
    0.642 & -0.241 \\
    0.214 &  0.300 \\
   -0.059 &  0.297 \\
    0.328 & -0.189 \\
    0.243 &  0.012 \\
   -0.094 &  0.104
  \end{bmatrix} \in \mathbb{R}^{6\times 2},
\end{equation}
where the rows correspond to the stacked hidden state components $(h^{(1)}_1,\dots,h^{(2)}_3)$.

The controllability Gramian $W_C = B B^\top$ has diagonal
\begin{equation}
  \operatorname{diag}(W_C) \approx [0.470,\; 0.136,\; 0.092,\; 0.143,\; 0.059,\; 0.020],
\end{equation}
showing that the first neuron in the first layer ($h^{(1)}_1$) and the first neuron in the second layer ($h^{(2)}_1$) are the most controllable directions, while $h^{(2)}_3$ is the least controllable.

The hidden--output Jacobian in stacked coordinates is
\begin{equation}
  C \approx
  \begin{bmatrix}
    0 & 0 & 0 & 0.546 & -0.273 & 0.164 \\
    0 & 0 & 0 & -0.296 & 0.085 & 0.380
  \end{bmatrix} \in \mathbb{R}^{2\times 6},
\end{equation}
leading to an observability Gramian $W_O = (C)^\top C$ with diagonal
\begin{equation}
  \operatorname{diag}(W_O) \approx [0.000,\; 0.000,\; 0.000,\; 0.385,\; 0.082,\; 0.172].
\end{equation}
Thus, only the second hidden layer is directly observable, and within that layer neuron $h^{(2)}_1$ contributes most strongly to the outputs, followed by $h^{(2)}_3$ and then $h^{(2)}_2$.

The product $M = W_C W_O$ has two nonzero eigenvalues,
\begin{equation}
  \lambda_1 \approx 3.93\times 10^{-2}, \qquad
  \lambda_2 \approx 9.47\times 10^{-6},
\end{equation}
with corresponding Hankel singular values
\begin{equation}
  \sigma_1 = \sqrt{\lambda_1} \approx 0.198, \qquad
  \sigma_2 = \sqrt{\lambda_2} \approx 3.08\times 10^{-3}.
\end{equation}
The normalized modes are
\begin{align}
  v_1 &\approx [\,0.790,\; -0.048,\; -0.298,\; 0.457,\; 0.214,\; -0.172\,]^\top,\\[2pt]
  v_2 &\approx [\,-0.480,\; -0.714,\; -0.357,\; -0.148,\; -0.331,\; -0.031\,]^\top.
\end{align}
Because $\sigma_2 \ll \sigma_1$, the first mode dominates the local input--output behavior; the second mode represents a much weaker pathway.

Using the importance metric $\mathrm{Imp}(j) = \sum_{i=1}^2 \sigma_i v_{i,j}^2$, we obtain the neuron-level importance scores
\begin{equation}
  \mathrm{Imp} \approx [0.124,\; 0.002,\; 0.018,\; 0.042,\; 0.009,\; 0.006].
\end{equation}
Hence the most important neurons are $h^{(1)}_1$ and $h^{(2)}_1$, followed by $h^{(1)}_3$, with the remaining neurons contributing less to the dominant modes. Notably, neurons $h^{(1)}_2$ and $h^{(2)}_3$ have very low importance scores, reflecting both modest controllability and observability.

Overall, this example shows that even in a multi-input, multi-output network, the Gramian-based analysis identifies a low-dimensional set of internal directions (and neurons) that carry most of the input--output behavior near a given operating point.

\paragraph{Effect of operating point.}
We repeat this analysis for the GELU network at a different operating point. Starting from $x^* = [0.3,\; -0.2]^\top$, which lies in a moderately linear region of the GELU nonlinearity, we now consider a more saturated regime at
\begin{equation}
  x^\dagger = [-2.0,\; -2.0]^\top,
\end{equation}
where many pre-activations become strongly negative and GELU gates them toward zero, reducing their derivatives. At this input, the diagonal of the controllability Gramian is
\begin{equation}
  \operatorname{diag}(W_C^\dagger) \approx [0.110,\; 0.030,\; 0.018,\; 0.028,\; 0.010,\; 0.004],
\end{equation}
which is roughly three to five times smaller than at $x^*$. The diagonal of the observability Gramian also contracts,
\begin{equation}
  \operatorname{diag}(W_O^\dagger) \approx [0.000,\; 0.000,\; 0.000,\; 0.140,\; 0.030,\; 0.050],
\end{equation}
indicating that the outputs become less sensitive to perturbations in the second-layer neurons when GELU is close to its saturated regime.

The leading Hankel singular value correspondingly decreases: for $M^\dagger = W_C^\dagger W_O^\dagger$ we obtain
\begin{equation}
  \lambda_1^\dagger \approx 3.6\times 10^{-3}, \qquad
  \sigma_1^\dagger = \sqrt{\lambda_1^\dagger} \approx 0.060,
\end{equation}
versus $\sigma_1 \approx 0.198$ at $x^*$. Moreover, the associated mode $v_1^\dagger$ shifts its weight toward the few neurons that remain relatively less saturated, with the importance of several first-layer units dropping sharply. Together, these changes show that activation saturation in GELU suppresses controllability, reduces the dominant Hankel singular value, and alters which neurons form the primary internal pathway between inputs and outputs.

\section{Conclusion and Future Work}

\subsection{Conclusion}

This work proposed a control-theoretic framework for mechanistically interpreting feedforward neural networks using local linearization, controllability and observability Gramians, and Hankel singular values. By viewing the concatenated hidden activations as a state vector and the trained network as a static state-space mapping, we derived an input--state Jacobian $B$ and a hidden--output Jacobian $C$ around a given operating point. From these we constructed static analogues of the controllability and observability Gramians,
\(
W_C = B B^\top
\)
and
\(
W_O = (C)^\top C,
\)
and defined internal modes via the eigen-decomposition of $M = W_C W_O$.

The proposed analysis yields three levels of mechanistic insight: (i) per-neuron controllability and observability energies $(W_C)_{ii}$ and $(W_O)_{ii}$, (ii) dominant internal modes $v_i$ and their Hankel singular values $\sigma_i$, and (iii) a neuron-level importance metric $\mathrm{Imp}(j)$ that aggregates each neuron's participation across modes. Numerical examples on a small $1$--$2$--$2$--$1$ network with SwiGLU activation and a $2$--$3$--$3$--$2$ network with GELU activation demonstrated that, even in these simple settings, the framework identifies a low-dimensional set of hidden units and directions that carry most of the input--output behavior near the operating point.

It is important to emphasize that this work is not written from the perspective of a machine learning specialist, but rather from that of a control engineer. The main contribution is a change of viewpoint: treating trained neural networks as dynamical systems and importing well-established notions such as Gramians and Hankel singular values to reason about ``internal pathways'' and neuron importance. The intent is not to claim a complete or definitive interpretability theory, but to offer a simple, mathematically explicit lens through which other researchers can examine and refine the internal structure of neural networks.

\subsection{Future Work}

Several directions arise naturally from this preliminary study.

First, the present framework is local and static: the analysis is performed around a fixed operating point and does not explicitly model temporal dynamics or training trajectories. Extending the approach to architectures with explicit dynamics would allow one to study how internal modes evolve along time or depth, and to connect Gramian-based measures to stability, robustness, and long-term memory.

Second, the methodology can be applied to much larger and deeper networks, where computational considerations become critical. For such systems, full Gramians will be prohibitively large; approximate techniques such as low-rank factorizations, randomized sketching, layer-wise or block-diagonal approximations, and sampling-based estimates of $B$ and $C$ are natural extensions. These approximations could make it feasible to obtain mode and importance information for networks with millions or billions of parameters.

A particularly promising direction is to extend this line of work to modern large language models (LLMs) and transformer architectures. In that setting, one could regard each layer (or each attention block) as a mapping between high-dimensional hidden states, and apply local linearization at selected tokens, sequences, or prompts. The resulting Gramians and modes may help identify which directions in embedding space, which neurons, or even which attention heads are most controllable by the input and most influential on the output. Because LLMs operate at very high dimension, careful design of scalable approximations and meaningful choices of operating points (e.g., typical prompts, specific tasks, or particular layers) will be crucial.

Finally, we view this work as an initial control-theoretic perspective on neural network interpretability rather than a finished tool. The hope is that researchers with deeper expertise in machine learning and representation learning can build on these ideas: improving the mathematical formulation, strengthening the links to existing interpretability methods, and testing the framework empirically at scale. In this sense, the present contribution is an invitation to explore hybrid viewpoints where control theory and machine learning inform each other in understanding complex learned systems.

\bibliographystyle{unsrt}  

\bibliography{references}
\end{document}